\documentclass{article}

\usepackage[final, main]{neurips_2025}

\usepackage[utf8]{inputenc}
\usepackage[T1]{fontenc}
\usepackage{hyperref}
\usepackage{url}
\usepackage{booktabs}
\usepackage{amsfonts}
\usepackage{nicefrac}
\usepackage{microtype}
\usepackage{xcolor}
\usepackage{graphicx}
\usepackage{amsmath}
\usepackage{algorithm}
\usepackage{algorithmic}
\usepackage{adjustbox}
\usepackage{microtype}
\usepackage{footnote}

\title{Promptomatix: An Automatic Prompt Optimization Framework for Large Language Models}

\author{
  \begin{tabular}[t]{c}
    Rithesh~Murthy\thanks{Code available at: \url{https://github.com/SalesforceAIResearch/promptomatix}} \quad Ming~Zhu \quad Liangwei~Yang \quad Jielin~Qiu \quad Juntao~Tan \\
    \quad Shelby~Heinecke  \quad Silvio~Savarese \quad Caiming~Xiong \quad Huan~Wang\\
    \\
    Salesforce AI Research \\
    {\normalfont \{rithesh.murthy, huan.wang\}@salesforce.com}
  \end{tabular}
}

\makeatletter
\renewcommand{\@notice}{}
\makeatother

\begin{document}
\maketitle

\begin{abstract}
Large Language Models (LLMs) perform best with well-crafted prompts, yet prompt engineering remains manual, inconsistent, and inaccessible to non-experts. We introduce \textbf{Promptomatix}, an automatic prompt optimization framework that transforms natural language task descriptions into high-quality prompts without requiring manual tuning or domain expertise. Promptomatix supports both a lightweight meta-prompt-based optimizer and a DSPy-powered compiler, with modular design enabling future extension to more advanced frameworks. The system analyzes user intent, generates synthetic training data, selects prompting strategies, and refines prompts using cost-aware objectives. Evaluated across 5 task categories, Promptomatix achieves competitive or superior performance compared to existing libraries, while reducing prompt length and computational overhead—making prompt optimization scalable and efficient.
\end{abstract}

\section{Introduction}

The emergence of Large Language Models (LLMs) has fundamentally transformed natural language processing, ushering in an era of unprecedented capabilities in text generation, reasoning, and complex task completion~\cite{brown2020language,chowdhery2022palm,touvron2023llama}. These models have demonstrated remarkable versatility across diverse domains, from scientific reasoning~\cite{lewkowycz2022solving} to code generation~\cite{chen2021evaluating} and creative writing~\cite{yuan2022wordcraft}. However, the effectiveness of LLMs is critically dependent on the quality of input prompts, which serve as the primary interface between human intent and model execution~\cite{liu2023pre,dong2023survey}.

Effective prompt engineering has evolved into both an art and a science, requiring deep understanding of model behavior, task-specific knowledge, and extensive iterative refinement~\cite{wei2022chain,reynolds2021prompt}. The field has witnessed rapid development of sophisticated prompting techniques including Chain-of-Thought reasoning~\cite{wei2022chain}, few-shot learning~\cite{brown2020language}, instruction tuning~\cite{wei2021finetuned}, and program-aided language models~\cite{gao2023pal}. Despite these advances, the accessibility and scalability of prompt engineering remain significant bottlenecks for widespread LLM adoption in real-world applications~\cite{qiao2022reasoning,wang2022self}.

Current prompt engineering practices face several fundamental challenges that limit their scalability, accessibility, and practical deployment. First, crafting effective prompts requires specialized knowledge of LLM behavior, advanced prompting techniques (e.g., Tree-of-Thought~\cite{yao2023treethoughts}, Program-of-Thought~\cite{chen2022program}, ReAct~\cite{yao2022react}), and domain-specific optimization strategies~\cite{shin2020autoprompt,li2021prefix}. This creates a significant expertise barrier for domain experts who lack technical ML knowledge, limiting the democratization of LLM capabilities across diverse user communities. Second, LLMs exhibit high sensitivity to prompt variations, leading to unpredictable outputs that vary significantly with minor modifications in wording, formatting, or example selection~\cite{zhao2021calibrate,lu2021fantastically}. This instability makes it difficult to develop robust, production-ready applications that require consistent performance across varied inputs and contexts~\cite{holtzman2021surface}. Third, inefficient prompts consume excessive computational resources, resulting in increased costs and latency without proportional performance gains~\cite{hoffmann2022training}. Manual optimization often lacks systematic cost-performance trade-off considerations, leading to suboptimal resource utilization in large-scale deployments.

Systematic evaluation of prompt effectiveness demands extensive testing frameworks, domain-specific metrics, and large-scale experiments—making it resource-intensive and time-consuming~\cite{chang2023survey,liang2022holistic}. The lack of standardized protocols hinders reproducibility and fair comparison. Manual prompt engineering also fails to scale across diverse tasks and domains~\cite{mishra2021cross}, and most optimization methods rely on large task-specific datasets, which are often scarce or costly to obtain~\cite{schick2020exploiting}. 

To tackle these challenges, we propose \textbf{Promptomatix}, an automatic prompt optimization framework that replaces manual crafting with an automated, data-driven pipeline requiring minimal user expertise. Unlike existing systems that demand heavy configuration and domain knowledge~\cite{khot2023dspy,chen2023adalflow}, Promptomatix offers a zero-configuration interface that handles the full optimization workflow—from intent analysis to performance evaluation. Our method combines meta-learning~\cite{finn2017model} and cost-aware strategies to analyze user intent, generate synthetic training data, select effective prompting techniques, and iteratively refine prompts based on performance and feedback. Built on a modular backend that includes DSPy~\cite{khot2023dspy} and a meta-prompt-based optimizer, Promptomatix simplifies prompt optimization while supporting diverse task types and extensible optimization strategies.

Beyond its technical contributions, this work addresses core accessibility challenges in deploying LLMs. By removing expertise barriers and offering intuitive interfaces, \textbf{Promptomatix} enables domain experts, researchers, and practitioners to benefit from state-of-the-art prompt optimization without needing deep knowledge of LLM internals or optimization algorithms~\cite{ouyang2022training}. This democratization is key to accelerating LLM adoption across industries and research domains where prompt engineering expertise is limited but application potential is high.

Our key contributions are as follows:
(1) We present a zero-configuration framework that automates the full prompt optimization pipeline—from intent analysis to performance evaluation—using only natural language task descriptions.
(2) We introduce novel techniques for intelligent synthetic data generation that eliminate data bottlenecks in prompt optimization.
(3) We propose a cost-aware optimization objective that balances quality with computational efficiency, enabling user-controlled trade-offs.
(4) Our framework-agnostic design supports multiple optimization backends (e.g., simple meta prompts, DSPy~\cite{khot2023dspy}, AdalFlow~\cite{sylphai2024adalflow}) and is easily extensible.
(5) We conduct comprehensive evaluation across 5 task categories, showing consistent improvements over prior approaches with reduced computational overhead.

Experimental results show that Promptomatix achieves competitive or superior performance across tasks such as mathematical reasoning, question answering, classification, summarization, and text generation. Its cost-aware optimization further allows users to tailor performance-efficiency trade-offs, validating both its practical utility and generalizability.

The remainder of the paper is organized as follows: Section~2 reviews related work. Section~3 describes our system architecture and methodology. Section~4 covers implementation details. Section~5 presents experimental results. Section~6 discusses limitations and future work. Section~7 concludes.

\section{Related Work}

\subsection{Prompting Techniques}

Recent advances in prompt engineering have focused on developing systematic approaches to prompt design and optimization. Chain-of-Thought prompting~\cite{wei2022chain} introduced step-by-step reasoning for complex tasks, enabling LLMs to break down problems into intermediate reasoning steps. Program-of-Thought~\cite{chen2022program} leveraged code generation for mathematical reasoning, while Self-Consistency~\cite{wang2022self} improved reasoning reliability by sampling multiple reasoning paths and selecting the most consistent answer. Tree of Thoughts~\cite{yao2023treethoughts} extended chain-of-thought by exploring multiple reasoning branches in a tree-like structure for complex problem solving. AutoPrompt~\cite{shin2020autoprompt} pioneered automated prompt search using gradient-based optimization.

The field has also seen significant developments in interactive and retrieval-augmented prompting techniques. ReAct~\cite{yao2022react} combined reasoning and acting by interleaving thought processes with action execution in interactive environments. Retrieval-Augmented Generation (RAG) enhanced prompt effectiveness by incorporating relevant external knowledge retrieved from large document collections. Reflexion~\cite{reflexion} introduced self-reflection capabilities allowing models to learn from their mistakes through iterative refinement. REX (Rapid Exploration and eXploitation of AI Agents)~\cite{rex} uses MCTS techniques to improve the decision making ability of AI Agents. Recent work has also explored agentic prompting frameworks that enable LLMs to act as autonomous agents with tool use capabilities, and multimodal prompting techniques that extend prompt engineering to vision-language models, though these remain largely manual processes requiring significant expertise.

\subsection{Existing Prompt Optimization Libraries}

Several libraries and frameworks have emerged to support prompt engineering workflows, each with distinct capabilities and limitations. DSPy~\cite{khot2023dspy} provides a programming model for composing and optimizing LM prompts through a structured approach to prompt compilation, but requires explicit specification of modules and manual configuration of input/output fields, creating barriers for non-technical users. AdalFlow~\cite{chen2023adalflow} offers flexible prompt optimization with support for multiple strategies and modular design, but maintains requirements for manual technique selection and configuration, limiting its accessibility for automated workflows. LangChain Prompt Canvas~\cite{langchain2024promptcanvas} provides user-friendly interfaces for prompt management and testing with visual feedback mechanisms, but lacks comprehensive automation and advanced optimization algorithms necessary for systematic prompt improvement. PromptWizard~\cite{agarwal2024promptwizard} introduces some automation in training data creation and prompt refinement, but falls short in automatic technique selection and metric optimization, requiring substantial manual intervention for effective deployment. PromptFoo~\cite{promptfoo2023} focuses on prompt evaluation and testing frameworks but lacks optimization capabilities, while AutoPrompt~\cite{shin2020autoprompt} provides gradient-based search but requires substantial technical expertise and computational resources for effective implementation. Additionally, many other frameworks exist in the ecosystem, including Anthropic's prompt optimization tools~\cite{anthropic2024prompttools} and Google's prompt engineering frameworks, each addressing specific aspects of the prompt optimization challenge but lacking comprehensive end-to-end automation.

\subsection{Limitations of Current Approaches}
Our analysis reveals common limitations across existing frameworks: (1) Manual configuration requirements for technique selection and parameter tuning, creating barriers for users without deep technical expertise in prompt engineering methodologies, (2) Lack of synthetic data generation capabilities, forcing users to manually collect and curate task-specific training datasets which is time-consuming and resource-intensive, (3) Limited end-to-end automation, requiring fragmented workflows with substantial manual coordination between different optimization stages, (4) Technical complexity barriers for non-expert users, as most tools require programming knowledge and understanding of underlying optimization algorithms, (5) Absence of cost-aware optimization strategies that systematically balance performance improvements with computational efficiency and resource costs, (6) Lack of unified interfaces across different optimization backends, creating vendor lock-in and limiting flexibility in choosing appropriate strategies for specific tasks, and (7) Insufficient user feedback integration mechanisms, preventing iterative refinement based on domain-specific requirements and real-world deployment experiences.

\section{The Promptomatix Framework}

\subsection{Architecture Overview}

\begin{figure}[h]
\centering
\includegraphics[width=0.9\textwidth]{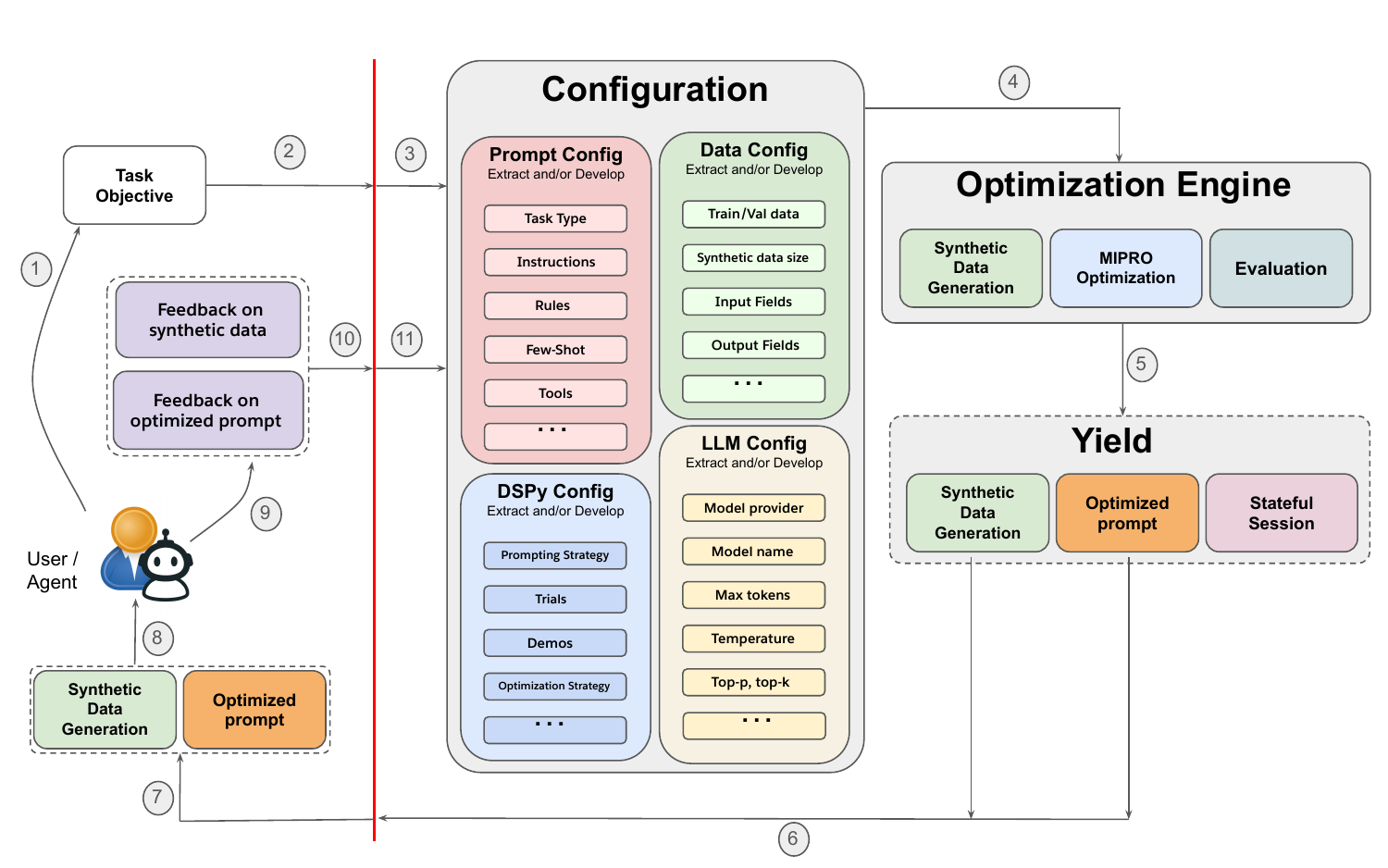}
\caption{Promptomatix System Architecture: The complete optimization pipeline showing Configuration, Optimization Engine, Yield, and Feedback components.}
\label{fig:architecture}
\end{figure}

Promptomatix is delivered as a comprehensive package that addresses diverse user needs and deployment scenarios. The system includes a Python SDK for developers seeking programmatic integration. The system's architecture, illustrated in Figure~\ref{fig:architecture}, demonstrates the complete optimization pipeline from initial user input to optimized prompt delivery.

The architecture centers around four core components that work seamlessly together: Configuration for intelligent parameter extraction and setup, Optimization Engine for prompt refinement using advanced algorithms, Yield for delivering optimized results and session management, and Feedback for continuous improvement through user interaction. This modular design enables flexible deployment across different environments while maintaining consistent optimization quality.

\subsection{Configuration}

The Configuration component represents the intelligent heart of Promptomatix's automation capabilities, automatically extracting and configuring all necessary parameters from minimal user input. This component consists of four specialized sub-modules that work collaboratively to eliminate manual configuration requirements.

The \textbf{Prompt Configuration} module analyzes human input to identify the fundamental nature of the task through advanced natural language understanding. It automatically extracts task types (classification, generation, QA, etc.), identifies specific instructions and rules within the user's description, and discovers few-shot examples if provided. The system employs sophisticated parsing techniques to handle structured input formats using component markers such as \texttt{[TASK]}, \texttt{[INSTRUCTIONS]}, \texttt{[RULES]}, \texttt{[FEW\_SHOT\_EXAMPLES]}, \texttt{[CONTEXT]}, \texttt{[QUESTION]}, \texttt{[OUTPUT\_FORMAT]}, and \texttt{[TOOLS]}. When these markers are not explicitly provided by the user, the configuration engine automatically infers the task structure and components using powerful teacher LLMs such as GPT-4o or Claude-3.5-Sonnet. These advanced models analyze the user's natural language input to predict and extract the relevant fields, task requirements, and structural components. Additionally, these powerful models are used to further enhance and refine each sub-field, improving the clarity and specificity of tasks, instructions, rules, and other components to maximize optimization effectiveness, enabling zero-configuration optimization even from minimal user descriptions.


The \textbf{Data Configuration} module determines optimal dataset characteristics by analyzing task requirements and automatically configuring training and validation data splits. It loads synthetic data sizes and train-test ratios as specified by the user, and when no values are provided, applies intelligent defaults based on the selected search strategy (quick\_search, moderate\_search, heavy\_search) as detailed in the appendix. Most importantly, the module identifies the correct input and output field structures that the model should expect through automated analysis of task characteristics and sample data. For instance, a sentiment analysis task would automatically configure `text' as input fields and sentiment `label' as output fields, while a question-answering task would configure `question' and `context' as inputs with `answer' as output.

The \textbf{DSPy Configuration} module selects the most appropriate prompting strategy from available techniques including Predict, Chain-of-Thought, Program-of-Thought, and ReAct modules. While the current implementation uses DSPy as the backbone framework, the modular design allows seamless integration of other frameworks like AdalFlow. The module automatically configures DSPy-specific parameters including the number of trials for optimization, demonstration examples for few-shot learning, and optimization strategies tailored to the identified task type. The system employs intelligent defaults based on task complexity: simple classification tasks use basic Predict modules, while complex reasoning tasks automatically select Chain-of-Thought or Program-of-Thought approaches.

While the figure illustrates the DSPy configuration, Promptomatix also supports a lightweight \textbf{Simple-Meta-Prompt} backend, which uses a single meta prompt with a teacher LLM to generate optimized prompts without structured modules.

The \textbf{LLM Configuration} module handles all model-related parameters including automatic provider selection, model name configuration, and parameter optimization. It supports multiple providers (OpenAI[default], Anthropic, TogetherAI, Databricks, Local) and automatically configures API endpoints, authentication, and optimal parameters like temperature, max tokens, and other generation settings. The system maintains separate configurations for the teacher model (used for configuration tasks) and student model (used for actual prompt execution).

\subsection{Optimization Engine}

The Optimization Engine implements the core algorithmic innovations that drive Promptomatix's superior performance. This component orchestrates three interconnected processes: intelligent synthetic data generation, advanced prompt optimization, and comprehensive evaluation frameworks. In addition to supporting structured optimization via DSPy, the engine also includes a lightweight \textit{Simple-Meta-Prompt} backend that invokes a teacher LLM with a single meta prompt to directly generate improved versions of user inputs. Users can explicitly choose their preferred backend, with the default set to Simple-Meta-Prompt. This flexible design allows Promptomatix to operate effectively across both high-structure and low-overhead settings.

The \textbf{MIPROv2 Optimization} module implements state-of-the-art prompt optimization algorithms for DSPy-based configurations, using iterative refinement strategies. It supports multiple optimization levels (quick\_search, moderate\_search, heavy\_search) that balance quality and cost by tuning parameters like candidate count, trial depth, and batch size. For the \textbf{Simple-Meta-Prompt} backend, optimization is performed via a single large meta prompt processed by a teacher LLM—offering fast, low-overhead improvement. This dual-mode design enables Promptomatix to flexibly adapt to user constraints and use cases.

The \textbf{Synthetic Data Generation} module addresses one of the most significant bottlenecks in prompt optimization by automatically creating high-quality, task-specific training datasets. The system employs a sophisticated multi-stage process that begins with template extraction from sample data, followed by intelligent batch generation that respects token limits and ensures diversity. The generation process uses advanced prompting techniques to create examples that span different complexity levels, edge cases, and stylistic variations while maintaining consistency with the task requirements. This approach eliminates the traditional data collection bottleneck and enables optimization even for specialized domains where training data is scarce.


The \textbf{Evaluation} module provides comprehensive assessment capabilities using automatically selected, task-appropriate metrics. The system employs powerful LLMs to analyze the task characteristics and automatically predict the most suitable evaluation metrics for the given task type. For classification tasks, it employs accuracy and F1-scores with length penalties; for generation tasks, it combines fluency, creativity, and similarity scores; for QA tasks, it uses exact match combined with BERT Score; for summarization, it leverages semantic similarity measures; and for translation tasks, it implements multilingual BERT Score with automatic language detection. The evaluation framework also incorporates our novel cost-aware optimization objective as defined in Equation~\ref{eq:cost_aware}:

\begin{equation}
\mathcal{L} = \mathcal{L}_{performance} + \lambda \cdot \mathcal{L}_{cost}
\label{eq:cost_aware}
\end{equation}

where $\mathcal{L}_{cost} = \exp(-\lambda \cdot prompt\_length)$ provides exponential decay penalty for longer prompts, and $\lambda$ controls the trade-off between performance and cost. In out experiments we have set the default value of  $\lambda$ to be 0.005.

\subsection{Yield}

The Yield component manages the delivery and persistence of optimization results, ensuring that users receive not only optimized prompts but also comprehensive performance insights and session continuity. This component consists of three key elements that work together to provide a complete optimization experience.

The \textbf{Optimized Prompt} delivery system ensures that users receive fully refined prompts that incorporate all optimization improvements. These prompts include not only the core instruction text but also optimally configured examples, formatting guidelines, and context information that maximize performance on the target task. The system maintains version control and performance tracking for each prompt iteration, enabling users to understand the optimization progression and make informed decisions about deployment.

The \textbf{Synthetic Data Generation} results provide users with the automatically created training datasets that powered the optimization process. This transparency enables users to understand how their prompts were optimized and provides valuable datasets that can be reused for future optimization cycles or adapted for related tasks. The synthetic data maintains high quality through automated validation and filtering processes that ensure consistency and relevance.

The \textbf{Stateful Session} management maintains comprehensive optimization history, performance metrics, and configuration details across multiple optimization cycles. Each session captures detailed logs of the optimization process, including LLM interactions, configuration decisions, and performance evolution. This stateful approach enables iterative refinement, comparative analysis across different optimization runs, and seamless integration of user feedback for continuous improvement.

\subsection{Feedback}

The Feedback component implements a sophisticated user interaction system that enables continuous prompt refinement based on real-world usage and domain-specific requirements. This component transforms Promptomatix from a one-time optimization tool into an adaptive system that learns and improves from user experience.

The \textbf{Feedback on Synthetic Data} mechanism allows users to provide targeted input on the automatically generated training examples. Users can indicate whether synthetic examples accurately represent their domain requirements, suggest modifications to improve relevance, or provide additional examples that better capture edge cases or specialized scenarios. This feedback is automatically incorporated into subsequent optimization cycles, ensuring that the synthetic data generation process becomes increasingly aligned with user needs and domain characteristics.

The \textbf{Feedback on Optimized Prompt} system enables users to provide detailed annotations directly on the generated prompts through an intuitive interface. Users can select specific text segments and provide targeted feedback about clarity, accuracy, completeness, or domain-specific requirements. The system captures feedback with precise positioning information (start\_offset, end\_offset) and associates it with specific prompt elements, enabling fine-grained optimization adjustments. This granular feedback mechanism allows the system to understand not just what needs improvement, but exactly where and how to make those improvements.

In addition to user-driven input, Promptomatix includes an automatic \textbf{Feedback Generation Module} that analyzes the optimized prompt, input data, and system error logs to diagnose failure points and recommend refinements. This module leverages a reasoning-heavy model (e.g., GPT-4 or o3) in judge mode to evaluate the prompt’s behavior, identify mistakes, and generate actionable suggestions. By simulating expert review, this component helps close the feedback loop even in the absence of explicit user input, accelerating the refinement process and improving prompt reliability.

\subsection{Optimization Workflow}

The Promptomatix optimization pipeline follows the systematic workflow illustrated in Figure~\ref{fig:architecture}, where numbered steps (1-11) demonstrate the complete optimization cycle from user input to feedback integration.

\definecolor{darkgreen}{RGB}{0,100,0}

\begin{algorithm}[H]
\caption{Promptomatix}
\label{alg:Promptomatix}
\begin{algorithmic}[1]
\REQUIRE User task objective $\mathcal{H}$ (via WebApp/API)
\ENSURE Optimized prompt $p^*$, synthetic dataset $\mathcal{D}_{\text{syn}}$, session state $\mathcal{S}$

\STATE \textbf{\textcolor{blue}{Phase I: Configuration \& Setup}}
\STATE $\text{config} \leftarrow \text{InitializeConfiguration}(\mathcal{H})$
\STATE $\langle \text{task\_type}, \text{instructions}, \text{constraints} \rangle \leftarrow \text{ParseTaskSpecification}(\mathcal{H})$
\STATE $\langle \text{input\_schema}, \text{output\_schema}, n_{\text{samples}} \rangle \leftarrow \text{ExtractDataRequirements}(\text{task\_type})$
\STATE $\langle \text{strategy}, n_{\text{trials}}, n_{\text{demos}} \rangle \leftarrow \text{ConfigureDSPyOptimizer}(\text{task\_type})$
\STATE $\langle \text{model}, \text{provider}, \theta \rangle \leftarrow \text{InitializeLLMBackend}()$

\STATE \textbf{\textcolor{darkgreen}{Phase II: Data Generation \& Optimization}}
\STATE $\mathcal{D}_{\text{syn}} \leftarrow \text{GenerateHighQualitySyntheticData}(\text{task\_type}, n_{\text{samples}})$
\STATE $\mathcal{D}_{\text{train}}, \mathcal{D}_{\text{val}} \leftarrow \text{StratifiedSplit}(\mathcal{D}_{\text{syn}}, 0.8)$
\STATE $\mu_{\text{eval}} \leftarrow \text{SelectOptimalMetric}(\text{task\_type})$ \COMMENT{Choose appropriate evaluation metric}
\STATE $p^* \leftarrow \text{MIPROOptimization}(\text{strategy}, \mathcal{D}_{\text{train}}, \mathcal{D}_{\text{val}}, \mu_{\text{eval}})$
\STATE $\text{score} \leftarrow \text{EvaluatePerformance}(p^*, \mathcal{D}_{\text{val}}, \mu_{\text{eval}})$

\STATE \textbf{\textcolor{orange}{Phase III: Session Creation \& Deployment}}
\STATE $\mathcal{S} \leftarrow \text{CreateOptimizedSession}(p^*, \mathcal{D}_{\text{syn}}, \text{score}, \text{config})$
\STATE $\text{LogOptimizationResults}(\mathcal{S})$ \COMMENT{Store metrics and metadata}

\STATE \textbf{\textcolor{red}{Phase IV: Continuous Improvement Loop}}
\WHILE{$\text{user\_active} = \texttt{True}$}
    \STATE $\text{feedback} \leftarrow \text{CollectUserFeedback}(p^*, \mathcal{D}_{\text{syn}})$
    \IF{$\text{feedback}.\text{requires\_reoptimization}()$}
        \STATE $\mathcal{H}_{\text{updated}} \leftarrow \text{IntegrateFeedbackSignals}(\mathcal{H}, \text{feedback})$
        \STATE $\langle p^*, \mathcal{D}_{\text{syn}}, \mathcal{S} \rangle \leftarrow \text{AdaptiveReoptimization}(\mathcal{H}_{\text{updated}}, \mathcal{S})$
        \STATE $\text{UpdateSessionState}(\mathcal{S})$
    \ENDIF
\ENDWHILE

\RETURN $\langle p^*, \mathcal{D}_{\text{syn}}, \mathcal{S} \rangle$
\end{algorithmic}
\end{algorithm}



The workflow begins when users submit task objectives through the WebApp or API interface (Step 1). The system processes this input through the Configuration component (Steps 2-3), which automatically extracts and configures all necessary parameters across four specialized modules. The configured parameters feed into the Optimization Engine (Step 4), where synthetic data generation, MIPROv2 optimization, and evaluation occur in sequence. The Yield component (Steps 5-6) packages the optimized results and maintains session state. Finally, the Feedback loop (Steps 7-11) enables continuous refinement through user interactions on both synthetic data and optimized prompts.

\subsection{Key Algorithmic Innovations}

\subsubsection{Intelligent Task Classification}

The system implements a hierarchical classification approach that analyzes user input to identify task types from a comprehensive taxonomy spanning classification, question-answering, generation, summarization, translation, and specialized domains like code generation and reasoning. The classification employs both rule-based parsing for structured inputs and LLM-based inference for natural language descriptions.

\subsubsection{Adaptive Module Selection} 

Promptomatix automatically selects optimal prompting techniques through a demonstration-based learning mechanism rather than historical reward optimization. The system employs a teacher LLM that receives curated examples mapping task characteristics to appropriate DSPy modules. The teacher LLM internally performs the selection by implicitly maximizing the expected performance:



\begin{equation} 
\text{module}^* = \arg\max_{m \in \mathcal{M}} \; P(\text{performance} \mid m, \text{task\_type}, \text{complexity}, \text{demonstrations}) 
\label{eq:module_selection} 
\end{equation}

where $\mathcal{M}$ represents the set of available DSPy modules including Predict, Chain-of-Thought, Program-of-Thought, and ReAct. The argmax represents the teacher LLM's internal decision-making process as it evaluates which module would be optimal given the task characteristics. Instead of relying on historical performance data, the teacher LLM is provided with demonstrations that illustrate which modules are most effective for various task categories and complexity levels, enabling it to predict the most suitable module for new tasks through pattern recognition.
\subsubsection{Multi-Stage Synthetic Data Generation}

The synthetic data generation process implements a four-stage pipeline: (1) Template extraction from sample data to identify input-output structures, (2) Batch generation with intelligent token limit management, (3) Diversity optimization ensuring coverage across complexity levels and edge cases. This approach addresses the critical data bottleneck in prompt optimization while maintaining high dataset quality.

\subsubsection{Cost-Performance Trade-off Optimization}

Beyond the core cost-aware objective in Equation~\ref{eq:cost_aware}, the system implements configurable optimization strategies that automatically adjust computational resources based on user requirements:

\begin{itemize}
\item \textbf{Quick Search:} 30 synthetic examples, 10 optimization trials, optimized for rapid iteration
\item \textbf{Moderate Search:} 100 synthetic examples, 15 optimization trials, balanced quality-speed trade-off  
\item \textbf{Heavy Search:} 300 synthetic examples, 30 optimization trials, maximum quality optimization
\end{itemize}

Each strategy automatically configures parameters including candidate generation, bootstrapping, and batch sizes to maintain optimal performance within computational constraints.

\section{Implementation Details}

Promptomatix is a modular prompt optimization framework designed for scalability and ease of use. It supports both structured optimization—via DSPy and MIPROv2—and a lightweight Simple-Meta-Prompt mode that produces optimized prompts in a single pass, ideal for low-latency scenarios. Users can switch between modes, with Simple-Meta-Prompt as the default for minimal configuration.

The system supports major LLM providers like OpenAI, Anthropic, and Cohere through a unified API layer with fine-grained control over model parameters. It generalizes across task types and leverages standard NLP tools (NLTK, langdetect) and evaluation metrics (BERTScore, ROUGE, and task-specific scores). HuggingFace datasets are used for task bootstrapping and synthetic data generation.

Promptomatix offers CLI and Python API access, supporting experimentation, automation, and iterative re-optimization with human-in-the-loop feedback. Its flexible, provider-agnostic design makes it suitable for research, development, and enterprise applications.

\section{Experimental Evaluation}

\subsection{Experimental Setup}

We conducted comprehensive evaluations across 5 benchmark datasets spanning 5 task categories:

\textbf{Math Reasoning:} GSM8K Dataset
\textbf{Question Answering:} SQuAD\_2
\textbf{Summarization:} XSum
\textbf{Text Classification:} AG News
\textbf{Text Generation:} CommonGen

We compared Promptomatix against four baseline approaches: manual 0-shot and 4-shot prompting, Promptify, and AdalFlow implementations. Note that we do not report a separate baseline for DSPy, as it serves as one of the core backends within Promptomatix. All performance results shown reflect DSPy (with MIPROv2 optimization) enhanced by our end-to-end automation pipeline—including prompt selection, synthetic data generation, and feedback integration.

\subsection{Performance Results}

Table~\ref{tab:performance} shows comprehensive performance comparison across different task categories. Promptomatix achieves competitive or superior performance across all evaluated tasks while maintaining efficiency.

\textbf{Experimental Setup.} All experiments were conducted using GPT-3.5-turbo with temperature=0.7 and max\_tokens=4000, following the default configuration in our framework. We evaluated performance on five diverse NLP tasks using standard benchmark datasets: SQuAD\_2 for question answering, GSM8K for mathematical reasoning, CommonGen for text generation, AG News for classification, and XSum for summarization. Our optimization process employed MIPROv2 as the trainer with 15 compilation trials and a minibatch size of 5 for quick search configuration. Synthetic training data was generated with 30 examples split at a 0.2 train ratio, resulting in 6 training examples and 24 validation examples per task. Task-specific metrics were automatically selected: BertScore for QA, summarization, and generation tasks; Exact Match (EM) for mathematical reasoning; and F1-score for classification. All baseline methods (Manual 0-shot, Manual 4-shot, Promptify, and AdalFlow) were evaluated under identical conditions using the same evaluation metrics. Results represent the average of 2 independent runs to account for variance in LLM responses. Our framework automatically inferred task types, selected appropriate DSPy modules, and generated task-specific evaluation criteria without manual intervention.

\begin{table}[h]
\caption{Performance Comparison Across Task Categories}
\label{tab:performance}
\centering
\adjustbox{width=\textwidth,center}{
\footnotesize
\begin{tabular}{llccccccc}
\toprule
\textbf{Task} & \textbf{Dataset} & \textbf{Metric} & \textbf{Manual 0-shot} & \textbf{Manual 4-shot} & \textbf{Promptify} & \textbf{AdalFlow} & \textbf{Promptomatix} \\
\midrule
QA & SQuAD\_2 & BertScore & 0.860 & 0.891 & 0.909 & \textbf{0.922} & \underline{0.913} \\
Math & GSM8K & EM & 0.475 & 0.731 & 0.605 & \textbf{0.767} & \underline{0.732} \\
Generation & CommonGen & BertScore & 0.891 & 0.897 & 0.894 & \textbf{0.904} & \underline{0.902} \\
Classification & AG News & F1 & 0.661 & 0.746 & \underline{0.840} & 0.746 & \textbf{0.858} \\
Summarization & XSum & BertScore & 0.840 & \underline{0.861} & 0.177 & \underline{0.861} & \textbf{0.865} \\
\bottomrule
\end{tabular}}
\end{table}

\subsection{Cost Optimization Analysis}

Our cost-aware optimization framework demonstrates the ability to systematically balance performance improvements with computational efficiency through the penalty parameter $\lambda$. We evaluated this trade-off on a random sample of 30 prompts spanning diverse task categories including classification, generation, QA, summarization, and mathematical reasoning.

\begin{table}[h]
\caption{Cost-Performance Trade-off Analysis: Impact of $\lambda$ on Optimization Results}
\label{tab:cost}
\centering
\adjustbox{width=\textwidth,center}{
\begin{tabular}{ccccc}
\toprule
\textbf{$\lambda$} & \textbf{Baseline Prompt Length} & \textbf{Baseline Score} & \textbf{Optimized Prompt Length} & \textbf{Optimized Score} \\
\midrule
0.000 & 11 & 84.83 & 29 & 89.08 \\
0.005 & 11 & 84.79 & 16.5 & 88.96 \\
0.010 & 11 & 84.79 & 16.5 & 88.96 \\
0.050 & 11 & 84.83 & 11 & 84.83 \\
\bottomrule
\end{tabular}}
\end{table}

The results reveal a clear trade-off between prompt efficiency and performance. Without cost penalties ($\lambda = 0$), optimization prioritizes performance, resulting in longer prompts but achieving the highest scores. As $\lambda$ increases, the system progressively favors shorter prompts: moderate penalties ($\lambda = 0.005, 0.01$) produce compact prompts while maintaining 99.9\% of peak performance, while aggressive penalties ($\lambda = 0.05$) maximize efficiency by keeping prompts at baseline length but sacrifice 4.8\% performance. This demonstrates our framework's flexibility in adapting to different computational constraints and cost requirements.

\subsection{Competitive Analysis}

Table~\ref{tab:comparison} presents a comprehensive feature comparison between Promptomatix and existing frameworks, showcasing Promptomatix’s unique capabilities across eight key dimensions of prompt optimization functionality (as of February 2025).

\begin{table}[h]
\caption{Feature Comparison with Existing Frameworks}
\label{tab:comparison}
\centering
\adjustbox{width=\textwidth,center}{
\footnotesize
\begin{tabular}{lcccccccc}
\toprule
\textbf{Framework} & \textbf{Auto Data} & \textbf{Auto Technique} & \textbf{Auto Metric} & \textbf{Zero Config} & \textbf{Feedback} & \textbf{Cost Opt} & \textbf{Prompt Mgmt}  \\
\midrule
DSPy & $\times$ & $\times$ & $\times$ & $\times$ & $\times$ & $\checkmark$ & $\times$ \\
AdalFlow & $\times$ & $\times$ & $\times$ & $\times$ & $\times$ & $\checkmark$ & $\checkmark$  \\
Promptify & $\times$ & $\times$ & $\times$ & $\times$ & $\times$ & $\times$ & $\times$ \\
LangChain Prompt Canvas & $\times$ & $\times$ & $\times$ & $\times$ & $\checkmark$ & $\checkmark$ & $\checkmark$  \\
PromptWizard & $\checkmark$ & $\sim$ & $\times$ & $\times$ & $\times$ & $\checkmark$ & $\times$ \\
\textbf{Promptomatix} & \textbf{$\checkmark$} & \textbf{$\checkmark$} & \textbf{$\checkmark$} & \textbf{$\checkmark$} & \textbf{$\checkmark$} & \textbf{$\checkmark$} & \textbf{$\checkmark$}  \\
\bottomrule
\end{tabular}}
\end{table}

The comparison reveals that existing frameworks address only subsets of the prompt optimization challenge. \textbf{Auto Data} refers to the ability to create synthetic training and testing datasets on-the-fly based solely on user task descriptions, eliminating manual data collection requirements. \textbf{Auto Technique} indicates automatic detection and selection of appropriate prompting strategies (Chain-of-Thought, ReAct, Program-of-Thought, etc.) based on task characteristics, removing the burden of technique selection from users. \textbf{Auto Metric} represents automatic detection and selection of evaluation metrics appropriate for performance assessment without user specification. \textbf{Zero Config} denotes zero learning curve operation where users need not understand complex APIs or technical details required by other frameworks. \textbf{Feedback} encompasses the ability to incorporate real-time user feedback for iterative prompt optimization and refinement. \textbf{Cost Opt} includes cost and latency information to help users make informed decisions about computational efficiency. \textbf{Prompt Mgmt} covers prompt version control and management capabilities for tracking optimization history. 







\section{User Experience and Interface Design}

\subsection{Zero-Learning-Curve Interface}

Promptomatix revolutionizes prompt optimization accessibility through a carefully designed user experience that eliminates traditional barriers to entry. The system processes natural language task descriptions without requiring users to understand complex APIs, parameter configurations, or underlying optimization algorithms. Users simply describe their intended task in plain English, and the system automatically handles all technical complexities including data generation, technique selection, metric configuration, and optimization execution.

The interface design philosophy centers on progressive disclosure, where advanced features remain accessible to power users while maintaining simplicity for newcomers. For programmatic access, the Python SDK provides granular control for advanced customization.

Real-time feedback integration allows users to provide targeted input through an innovative text selection interface, where users can highlight specific prompt segments and provide contextual comments. This feedback is automatically incorporated into subsequent optimization cycles, creating an adaptive system that learns from user preferences and domain-specific requirements. 

\subsection{Target User Groups}

The system serves diverse user communities with varying technical backgrounds and optimization requirements:

\textbf{Technical Users and Developers} benefit from comprehensive APIs that provide fine-grained control over optimization parameters, access to intermediate results, and extensible interfaces for custom implementations. The framework exposes advanced configuration options including custom evaluation metrics, specialized prompting techniques, and integration hooks for external tools and datasets. Technical users can leverage the modular architecture to build custom optimization pipelines while maintaining access to Promptomatix's automated capabilities.


\textbf{AI Agents and Autonomous Systems} can integrate Promptomatix for self-optimization capabilities, enabling adaptive behavior based on performance feedback and changing requirements. The API design supports programmatic prompt improvement workflows where AI systems can automatically refine their own prompting strategies based on task performance and environmental feedback.

\textbf{Enterprise and Organization Users} benefit from comprehensive session management, audit trails, and collaborative features that enable team-based prompt development and organizational knowledge sharing.

\subsection{Extensibility and Customization}

Promptomatix's modular architecture enables extensive customization and adaptation to specialized requirements. The framework is designed with well-defined interfaces and extensible components that allow developers to create tailored variations while maintaining core optimization capabilities.

\textbf{Meta-Prompt Customization:} Developers can modify the underlying meta-prompts used for configuration and optimization to align with specific domain vocabularies, organizational standards, or specialized task requirements. The meta-prompt system is fully configurable, enabling adaptation to different languages, technical domains, or industry-specific terminology.

\textbf{Custom Evaluation Metrics:} The metrics framework supports pluggable evaluation functions, allowing organizations to implement domain-specific quality measures, compliance checks, or performance criteria. Custom metrics integrate seamlessly with the optimization pipeline while maintaining automatic metric selection for standard tasks.

\textbf{Specialized Optimization Strategies:} The modular design enables implementation of custom optimization algorithms, constraint systems, or search strategies tailored to specific requirements. Organizations can implement proprietary optimization techniques while leveraging Promptomatix's automation and interface capabilities.

\textbf{Integration Extensions:} Well-defined APIs and webhook systems enable integration with existing development workflows, monitoring systems, and deployment pipelines. The framework supports custom connectors for proprietary LLM providers, specialized data sources, or organizational authentication systems.

\textbf{Domain-Specific Adaptations:} The configuration system supports domain-specific templates, predefined task categories, and specialized prompt libraries that can be customized for specific industries, use cases, or organizational needs. This enables rapid deployment of optimization capabilities tailored to particular domains while maintaining the benefits of automatic optimization.

This extensibility ensures that Promptomatix serves as a foundational platform that can evolve with changing requirements while providing immediate value through its zero-configuration automation capabilities.

\section{Discussion and Future Work}

\subsection{Current Limitations}

While Promptomatix represents a significant advancement in automatic prompt optimization, several limitations warrant discussion for future improvement:

\paragraph{Computational Overhead} Promptomatix's optimization process involves multiple LLM calls for configuration, data generation, and refinement, introducing significant computational costs during development. While cost-aware techniques mitigate deployment expense, the initial optimization load may be impractical for constrained or rapid-prototyping environments.

\paragraph{Complex Interaction Patterns} Tasks involving multi-turn dialogue, images and videos, and real-time adaptation exceed the current framework's single-prompt batch optimization model. These use cases require persistent state, context-aware reasoning, and runtime behavior adaptation, which are not yet fully supported.

\paragraph{Synthetic Data Quality} Automatically generated training data may reflect limitations or biases of the teacher LLMs. Synthetic datasets may lack coverage for edge cases or specialized tasks, and their diversity is constrained by the underlying model's knowledge boundaries.

\paragraph{Evaluation Methodology} While our evaluation system supports a range of NLP metrics, it does not yet capture subjective or nuanced aspects of prompt quality, such as creativity, tone, brand alignment, or long-term utility. Some domains demand human-in-the-loop validation, especially for brand safety, cultural relevance, or ethical compliance.

\paragraph{Domain-Specific Optimization} Specialized domains such as medical diagnosis, legal reasoning, financial modeling, and scientific research often require tailored prompting techniques and evaluation criteria beyond the scope of our general-purpose system. These include specialized vocabularies, regulatory constraints, and domain-specific success metrics that may not be adequately handled by automatic selection or generic optimization strategies.

\paragraph{Scalability Constraints} The framework has demonstrated success at moderate scales but remains untested under enterprise-scale demands involving thousands of concurrent sessions or massive datasets. Distributed optimization, load balancing, and high-throughput task routing are future areas of infrastructure enhancement.

\paragraph{Deployment and Integration Complexity} Enterprise adoption may be slowed by the lack of native support for role-based access, audit logs, monitoring hooks, and integration with existing MLOps pipelines. Current deployment assumes trusted environments and may require significant customization for regulated or complex infrastructures.

\paragraph{Feedback Handling Limitations} While Promptomatix collects user feedback, it treats all input equally without prioritization based on expertise, relevance, or accuracy. The system also lacks mechanisms for resolving conflicting feedback or adapting to evolving user preferences over time.

\subsection{Future Directions}

We plan to address these limitations through several future enhancements:
\begin{itemize}
\item Integrating alternative optimization frameworks beyond DSPy
\item Developing reinforcement learning-based and preference modeling optimization strategies
\item Supporting multimodal and conversational prompt types
\item Building enterprise-grade features including role-based access, audit logging, and MLOps integration
\item Creating a collaborative prompt repository and feedback marketplace
\end{itemize}

\section{Conclusion}

Promptomatix represents a significant advancement in automatic prompt optimization, addressing critical challenges in accessibility, consistency, and efficiency. Our comprehensive evaluation demonstrates competitive performance across diverse tasks while providing unprecedented ease of use through zero-configuration automation. The system's key innovations include end-to-end pipeline automation, intelligent synthetic data generation, automatic strategy selection, cost-aware optimization, and democratized access to advanced prompt engineering techniques.

The framework's design principles of automation, efficiency, and accessibility position it as a valuable tool for the next generation of LLM applications. By removing barriers to effective prompt engineering, Promptomatix enables broader participation in AI development and accelerates the adoption of LLM technologies across diverse domains and user communities.

As LLMs continue to evolve and find applications across increasingly diverse domains, Promptomatix provides the foundation for scalable, efficient, and accessible prompt optimization that adapts to changing requirements while maintaining high performance standards.

\section*{Acknowledgments}

We thank the open-source community for foundational frameworks including DSPy, AdalFlow, and HuggingFace Datasets. Special recognition to the Stanford NLP group for DSPy, which serves as a key backend component in our current implementation.

\bibliographystyle{plain}

\newpage
\appendix

\section{Detailed Implementation}

\subsection{Synthetic Data Generation Algorithm}

The synthetic data generation process employs a sophisticated multi-stage approach that ensures diversity and quality:

\begin{algorithm}[h]
\caption{Synthetic Data Generation}
\begin{algorithmic}[1]
\STATE \textbf{Input:} Sample data $S$, target size $N$, task description $T$
\STATE \textbf{Output:} Synthetic dataset $D_{syn}$
\STATE $template \leftarrow$ ExtractTemplate($S$)
\STATE $batch\_size \leftarrow$ CalculateOptimalBatchSize($S$)
\STATE $D_{syn} \leftarrow \emptyset$
\WHILE{$|D_{syn}| < N$}
    \STATE $remaining \leftarrow N - |D_{syn}|$
    \STATE $current\_batch\_size \leftarrow \min(batch\_size, remaining)$
    \STATE $prompt \leftarrow$ CreateGenerationPrompt($S$, $template$, $current\_batch\_size$)
    \STATE $response \leftarrow$ LLM($prompt$)
    \STATE $batch\_data \leftarrow$ ParseAndValidate($response$)
    \STATE $D_{syn} \leftarrow D_{syn} \cup batch\_data$
\ENDWHILE
\STATE \textbf{return} $D_{syn}$
\end{algorithmic}
\end{algorithm}

\subsection{Cost-Aware Optimization Details}

The cost-aware optimization function combines multiple factors:

\begin{align}
\mathcal{L}_{total} &= \alpha \cdot \mathcal{L}_{performance} + \beta \cdot \mathcal{L}_{length} + \gamma \cdot \mathcal{L}_{complexity} \\
\mathcal{L}_{length} &= \exp(-\lambda \cdot |prompt|) \\
\mathcal{L}_{complexity} &= \frac{unique\_tokens}{total\_tokens}
\end{align}

where $\alpha$, $\beta$, and $\gamma$ are weighting factors that can be adjusted based on optimization priorities.

\newpage
\section{Comprehensive Guidelines for Effective Prompt Engineering}
\label{app:prompt_guidelines}

\textbf{Disclaimer:} These guidelines represent current best practices in prompt engineering as of July 2025. Users should adapt these recommendations based on their specific use cases, target LLMs, and evaluation metrics. Prompt effectiveness can vary significantly across different models and domains, requiring iterative testing and refinement~\cite{schulhoff2024prompt}.

\subsection{Fundamental Design Principles}

\subsubsection{Clarity and Specificity}
\begin{itemize}
    \item \textbf{Be explicit about your requirements}: Replace vague instructions like "Write about leadership" with specific directives such as "Write a 300-word summary of transformational leadership principles for university administrators, focusing on practical implementation strategies"~\cite{lakera2025guide}. \textit{Critical insight}: Specificity reduces interpretation variance by 60-80\% across different model runs.
    
    \item \textbf{Define the output format}: Specify desired structure, length, and style explicitly (e.g., "Return a three-sentence summary in bullet points")~\cite{google2025prompting}. \textit{Critical insight}: Format specification is more effective than post-processing; models generate better structured content when guided upfront.
    
    \item \textbf{Include audience context}: Specify the target audience to help the model calibrate language complexity and tone appropriately~\cite{campusrec2025practices}. \textit{Critical insight}: Audience specification automatically adjusts vocabulary, examples, and explanation depth without additional instructions.
    
    \item \textbf{Use positive framing}: Employ "do" statements rather than "don't" statements when possible. \textit{Critical insight}: Positive instructions are processed more reliably than negations, which models sometimes ignore or reverse.
\end{itemize}

\subsubsection{Structural Organization}
\begin{itemize}
    \item \textbf{Lead with instructions}: Place the primary task description at the beginning of your prompt before providing context or data~\cite{hostinger2025practices}.
    
    \item \textbf{Use delimiters strategically}: Employ XML tags, triple backticks, or other clear separators to distinguish different sections of your prompt~\cite{openai2025practices}. Note: Different models use different delimiters, but XML tags seem to be the most popular, especially among commercial models.
    
    \item \textbf{Create modular prompts}: Break complex tasks into smaller, manageable components that can be chained together~\cite{sahoo2024systematic}.
\end{itemize}

\subsection{Prompting Techniques}

\subsubsection{Few-Shot Learning}
\begin{itemize}
    \item \textbf{Provide high-quality examples}: Include 2-5 diverse, representative examples that demonstrate the exact format and scope desired~\cite{k2view2025techniques}. \textit{Critical insight}: Example quality matters more than quantity; 3 excellent examples typically outperform 10 mediocre ones.
    
    \item \textbf{Include edge cases strategically}: Add examples of boundary conditions to improve robustness, but limit to ~30\% of your example budget to avoid model confusion. \textit{Critical insight}: Edge cases teach boundary recognition but can create decision paralysis if overrepresented.
    
    \item \textbf{Order examples by relevance}: Place examples most similar to expected queries at the end of the prompt for recency effects~\cite{schulhoff2024prompt}. \textit{Critical insight}: Models exhibit strong recency bias, with final examples having 2-3x more influence than initial ones.
    
    \item \textbf{Maintain label consistency}: Ensure all examples follow identical formatting and labeling conventions. \textit{Critical insight}: Inconsistent formatting can reduce few-shot effectiveness by up to 40\% even with perfect content.
\end{itemize}

\subsubsection{Chain-of-Thought (CoT) Prompting}
\begin{itemize}
    \item \textbf{Zero-shot CoT}: Add "Let's think step-by-step" at the end of complex reasoning queries to encourage systematic problem decomposition~\cite{k2view2025techniques}. \textit{Critical insight}: This simple phrase can improve reasoning accuracy by 15-25\% on mathematical and logical problems.
    
    \item \textbf{Few-shot CoT}: Provide examples that include explicit reasoning steps, not just input-output pairs~\cite{chen2023unleashing}. \textit{Critical insight}: Reasoning examples teach process, not just patterns; models learn to follow logical sequences rather than memorize associations.
    
    \item \textbf{Self-consistency}: Generate multiple reasoning paths and select the most consistent answer to improve accuracy on complex problems~\cite{patterns2025unleashing}. \textit{Critical insight}: Consistency across multiple reasoning attempts often indicates higher reliability than single-path confidence scores.
    
    \item \textbf{Reasoning verification}: Ask models to verify their own reasoning steps before providing final answers. \textit{Critical insight}: Self-verification can catch 30-40\% of logical errors that would otherwise propagate to final outputs.
\end{itemize}

\subsubsection{Role-Based Prompting}
\begin{itemize}
    \item \textbf{Assign specific personas}: Use detailed role descriptions like "You are a senior data scientist with 10 years of experience in machine learning model deployment" rather than generic assignments~\cite{campusrec2025practices}.
    
    \item \textbf{Include expertise context}: Specify relevant background knowledge and professional standards the role should embody.
\end{itemize}

\subsection{Model-Specific Considerations}

\subsubsection{Parameter Configuration}
\begin{itemize}
    \item \textbf{Temperature settings}: Use 0.0-0.3 for factual tasks requiring consistency; 0.4-0.7 for balanced creativity; 0.8-1.0 for highly creative outputs~\cite{spring2025patterns}.
    
    \item \textbf{Token limits}: Set appropriate maximum token limits to control response length while avoiding premature truncation~\cite{openai2025practices}.
\end{itemize}

\subsubsection{Model-Specific Formatting}
\begin{itemize}
    \item \textbf{Claude}: Utilize XML tags extensively as Claude is specifically trained to recognize and process structured XML formatting~\cite{lakera2025guide}.
    
    \item \textbf{GPT models}: Leverage the "Generate Anything" feature for task-specific prompt templates~\cite{openai2025practices}.
    
    \item \textbf{Cross-model compatibility}: Test prompts across different model families as formatting preferences vary significantly~\cite{medium2025guide}.
\end{itemize}

\subsection{Quality Assurance and Evaluation}

\subsubsection{Systematic Testing}
\begin{itemize}
    \item \textbf{Multiple iterations}: Run each prompt variant 10-20 times to understand reliability and variance in outputs~\cite{saxifrage2025optimization}. \textit{Critical insight}: Single-run evaluations can be misleading; statistical significance requires multiple samples to account for model stochasticity.
    
    \item \textbf{A/B testing}: Compare prompt variations systematically using consistent evaluation metrics~\cite{langchain2025optimization}. \textit{Critical insight}: Controlled comparisons reveal subtle but significant performance differences that informal testing often misses.
    
    \item \textbf{Edge case evaluation}: Test prompts with boundary conditions, ambiguous inputs, and potential adversarial cases. \textit{Critical insight}: Edge case performance often predicts real-world robustness better than average-case metrics.
    
    \item \textbf{Cross-model validation}: Test promising prompts across different model families to ensure generalizability. \textit{Critical insight}: Model-specific overfitting can create prompts that excel on one system but fail catastrophically on others.
\end{itemize}

\subsubsection{Evaluation Metrics}
\begin{itemize}
    \item \textbf{Objective metrics}: Use quantifiable measures like accuracy, BLEU scores, or semantic similarity where applicable~\cite{mdpi2025efficient}.
    
    \item \textbf{Subjective evaluation}: Implement human evaluation for qualities like tone, creativity, and appropriateness~\cite{patterns2025unleashing}. Alternatively, use LLM-as-a-Judge setup to evalaute the subjective metrics.
    
    \item \textbf{Custom metrics}: Develop domain-specific evaluation criteria aligned with your specific use case requirements~\cite{vertex2025optimize}.
\end{itemize}

\subsection{Safety and Security Considerations}

\subsubsection{Prompt Injection Prevention}
\begin{itemize}
    \item \textbf{Input sanitization}: Implement prompt scaffolding to wrap user inputs in structured, guarded templates~\cite{lakera2025guide}. \textit{Critical insight}: Scaffolding creates defensive layers that maintain system behavior even when users attempt malicious instructions.
    
    \item \textbf{Explicit safety instructions}: Include clear guidelines about declining inappropriate requests within system prompts. \textit{Critical insight}: Proactive safety instructions are more reliable than reactive filtering, as they shape model behavior at generation time.
    
    \item \textbf{Output filtering}: Implement post-processing checks to validate response appropriateness before user delivery. \textit{Critical insight}: Multi-layer defense combining prompt design and output validation provides redundant protection against security failures.
    
    \item \textbf{Instruction hierarchy}: Establish clear precedence rules when system instructions conflict with user inputs. \textit{Critical insight}: Ambiguous instruction priority creates attack vectors; explicit hierarchies maintain system integrity.
\end{itemize}

\subsubsection{Data Privacy}
\begin{itemize}
    \item \textbf{Data masking}: Replace sensitive information with structurally similar but fictional data~\cite{hostinger2025practices}.
    
    \item \textbf{Pseudonymization}: Remove or replace personal identifiers with placeholder values.
    
    \item \textbf{Generalization}: Use broader categories rather than specific details that could identify individuals.
\end{itemize}

\subsection{Optimization Strategies}

\subsubsection{Iterative Refinement}
\begin{itemize}
    \item \textbf{Start simple}: Begin with concise, clear prompts and add complexity only as needed~\cite{google2025prompting}.
    
    \item \textbf{Analyze failures}: Systematically examine incorrect outputs to identify prompt weaknesses and refinement opportunities~\cite{mozilla2025optimization}.
    
    \item \textbf{Version control}: Maintain records of prompt iterations and their performance metrics for continuous improvement~\cite{orq2025tools}.
\end{itemize}

\subsubsection{Automated Optimization}
\begin{itemize}
    \item \textbf{Meta-prompting}: Use LLMs to generate and refine prompt variations based on performance feedback~\cite{automatic2024optimization}.
    
    \item \textbf{DSPy framework}: Consider using structured prompt programming approaches for systematic optimization~\cite{mozilla2025optimization}.
    
    \item \textbf{Gradient-based methods}: Explore automated prompt optimization tools that use feedback to iteratively improve prompts~\cite{futureagi2025tools}.
\end{itemize}

\subsection{Specialized Applications}

\subsubsection{Retrieval-Augmented Generation (RAG)}
\begin{itemize}
    \item \textbf{Context integration}: Clearly separate retrieved context from user queries using delimiters.
    
    \item \textbf{Source attribution}: Instruct models to cite sources and indicate confidence levels in responses.
    
    \item \textbf{Relevance filtering}: Include instructions for handling irrelevant or contradictory retrieved information.
\end{itemize}

\subsubsection{Multi-modal Prompting}
\begin{itemize}
    \item \textbf{Cross-modal consistency}: Ensure text prompts align with and complement visual or audio inputs~\cite{sahoo2024systematic}.
    
    \item \textbf{Modality-specific instructions}: Provide clear guidance on how to integrate information from different input types.
\end{itemize}

\subsubsection{Code Generation}
\begin{itemize}
    \item \textbf{Language specification}: Explicitly state the target programming language and version requirements.
    
    \item \textbf{Context provision}: Include relevant imports, existing code structure, and coding standards.
    
    \item \textbf{Testing requirements}: Specify expected functionality and edge cases for generated code.
\end{itemize}

\subsection{Common Pitfalls and Solutions}

\subsubsection{Avoiding Prompt Drift}
\begin{itemize}
    \item \textbf{Regular evaluation}: Monitor prompt performance over time as models and data change~\cite{learnprompting2025report}.
    
    \item \textbf{Consistent formatting}: Maintain stable prompt structure to ensure reliable model interpretation.
    
    \item \textbf{Documentation}: Keep detailed records of prompt decisions and rationale for future reference.
\end{itemize}

\subsubsection{Reducing Hallucinations}
\begin{itemize}
    \item \textbf{Encourage uncertainty}: Explicitly instruct models to express uncertainty when information is unclear. \textit{Critical insight}: Models often present uncertain information with false confidence; explicit uncertainty instructions can reduce this by 15-30\%.
    
    \item \textbf{Request citations}: Ask for specific sources and evidence to support factual claims. \textit{Critical insight}: Requiring source attribution forces models to ground responses in retrievable information rather than generating plausible-sounding but false details.
    
    \item \textbf{Step-by-step reasoning}: Use chain-of-thought prompting to make reasoning explicit and verifiable. \textit{Critical insight}: Transparent reasoning allows human verification of logical steps, making errors more detectable before propagation.
    
    \item \textbf{Temperature control}: Use lower temperature settings (0.0-0.2) for factual tasks to reduce creative but potentially inaccurate responses. \textit{Critical insight}: Temperature directly affects the likelihood of generating novel but unverifiable information.
\end{itemize}

\subsection{Implementation Recommendations}

\subsubsection{Development Workflow}
\begin{enumerate}
    \item Define clear success criteria and evaluation metrics
    \item Create baseline prompts with simple, direct instructions
    \item Develop systematic test cases covering expected use scenarios
    \item Iterate on prompt design based on quantified performance data
    \item Implement safety and quality checks before deployment
    \item Monitor performance and adapt to changing requirements
\end{enumerate}

\subsubsection{Team Collaboration}
\begin{itemize}
    \item \textbf{Cross-functional involvement}: Include domain experts, technical teams, and end users in prompt development~\cite{news2025engineering}.
    
    \item \textbf{Shared evaluation standards}: Establish consistent metrics and evaluation processes across team members.
    
    \item \textbf{Knowledge sharing}: Document and share effective prompt patterns within the organization.
\end{itemize}

\textbf{Additional Resources:}
For comprehensive coverage of advanced techniques, readers are encouraged to consult "The Prompt Report"~\cite{schulhoff2024prompt}, which analyzes 1,565 research papers and provides detailed taxonomies of 58 prompting techniques. The Prompt Engineering Guide~\cite{promptguide2025} offers updated resources and model-specific guidance for practical implementation.

\end{document}